# Behaviour-based Knowledge Systems:
# An Epigenetic Path from Behaviour to Knowledge

**Carlos Gershenson**
School of Cognitive and Computer Sciences
University of Sussex
Brighton, BN1 9QN, U. K.
C.Gershenson@sussex.ac.uk
http://www.cogs.sussex.ac.uk/users/carlos

## Abstract

In this paper we expose the theoretical background underlying our current research. This consists in the development of behaviour-based knowledge systems, for closing the gaps between behaviour-based and knowledge-based systems, and also between the understandings of the phenomena they model. We expose the requirements and stages for developing behaviour-based knowledge systems and discuss their limits. We believe that these are necessary conditions for the development of higher order cognitive capacities, in artificial and natural cognitive systems.

## 1. Introduction

In the field of artificial intelligence, knowledge-based systems (KBS) and behaviour-based systems (BBS) have modelled and simulated exhibitions of intelligence of different types, which we could call "cognitive" intelligence and "adaptive" intelligence, respectively. Broadly, and independently of their methodologies, BBS have modelled intelligence exhibited by animals adapting to their environment, while KBS have modelled "higher" cognition: reasoning, planning, and problem solving. Trying to understand how this higher cognition could evolve from adaptive behaviour, we propose the development of *behaviour-based knowledge systems* (BBKS). They are systems where an artificial creature is able to abstract and develop through its behaviour knowledge from *its* environment, and exploit this knowledge for having a favourable performance in *its* environment. BBKS relate the exhibitions of intelligence modelled by BBS and KBS, closing the gaps between them.

In order to develop these ideas, in the next section we expose *abstraction levels* (Gershenson, 2002a) in animal behaviour, which are useful for illustrating our goals. In Section 3 we present the steps we believe should be followed in order to develop and exhibit knowledge parting from adaptive behaviour. In Section 4 we note limits of BBKS, which are related to the limits of Epigenetic Robotics and Artificial Intelligence. We also briefly describe our current work, which consists in the implementation of a BBKS.

## 2. Abstraction Levels in Animal Behaviour

Abstraction levels (Gershenson, 2002a) *represent* simplicities and regularities in nature. Phenomena are easier to represent in our minds when they are simple. We can have an almost clear concept of them, and then we can try to understand complex phenomena in terms of our simple representations. We can recognize abstraction levels in atoms, molecules, cells, organisms, societies, ecosystems, planets, planetary systems, galaxies. An element of an abstraction level has a simple and regular behaviour, and it is because of this that can be easily observed and described. At least easier than the complexities that emerge from the interactions of several elements.

We can identify abstraction levels in animal behaviour (Gershenson, 2001, pp. 2-3), taking the definition of behaviour developed by Maturana and Varela: "behaviour is a description an observer makes of the changes in a system with respect to an environment with which the system interacts" (Maturana and Varela, 1987, p. 163). Our proposal is not a final categorization, but it is quite convenient for orienting our work, even when the borders between levels are fuzzy. The most elemental type of behaviour is **vegetative**, which can be seen as behaviours "by default" (such as breathing, metabolism, etc.). We can also distinguish **reflex** behaviours. These are action-response-based behaviours (such as reactions to pain). Stepping-up in complexity, we can identify

**reactive** behaviours, which depend strongly of an external stimulus, or a set or sequence of external stimuli (McFarland, 1981). Examples of these can be locomotion patterns. These behaviours (and the ones which follow) require an action selection process, whereas reflex behaviours are executed whenever the triggering stimulus is present. **Motivated** behaviours do not only depend on external stimuli (or the absence of a specific stimulus), but also on internal motivations. For example, "exploration for food" can be performed when there is the internal motivation "hunger". The previous types of behaviour have been modelled with behaviour-based systems (BBS) (*e.g.* Brooks, 1986; Beer, 1990; Maes, 1990; 1993; Hallam, Halperin and Hallam, 1994; González, 2000; Gershenson, 2001). **Reasoned** behaviours are the ones which are determined by manipulations of abstract concepts or representations. Preparing yourself for a trip would be an example. You would like to make plans, for which you would need to have abstract representations, and very probably a language (Clark, 1998), and to manipulate these representations. This manipulation can be considered as the *use* of a *logic*. This level has been modelled with knowledge-based systems (KBS) (*e.g.* Newell and Simon, 1972; Lenat and Feigenbaum, 1992). We could speculate about **conscious** behaviours, without entering the debate of the definition consciousness, just saying that they are behaviours that are determined by the individual's consciousness. We do not believe that there is an "ultimate" level of behaviour. We could, in theory, always find behaviours produced by mechanisms more and more complex. But for now we have enough trying to model behaviours less complex than reasoned ones. If we cannot clearly identify further levels, there is no sense in trying to model them. Figure 1 shows a diagram of the types of behaviours described above.

We believe that the behaviours in the higher levels evolved and developed from the behaviours in the lower levels, since in animals you cannot find higher levels of behaviour without the lower ones. Thus, higher levels of behaviour require the lower ones, in a similar way as children need to develop first lower stages in order to reach higher ones (Piaget, 1968). Also the higher types of behaviour in many cases can be seen as complex variants of the lower ones. Therefore, it is sensible to attempt to build artificial cognitive systems exhibiting adaptive behaviour of higher levels incrementally: in a bottom-up fashion (Gershenson, 2001:3). This does not mean that we cannot model any level separately. But the more levels we consider, the less-incomplete our models will be.

Historically, KBS were used first trying to model and simulate the intelligence found at the level of reasoned behaviours in a *synthetic* way (Steels, 1995; Verschure, 1998; Castelfranchi, 1998). This means that we build an artificial system in order to test our model, instead of contrasting our model directly with observations on the modelled system. The synthetic method allows us to contrast our theories with artificial systems, and in the case of intelligence and mind, theories are very hard to contrast with the natural systems. KBS have proven to be acceptable models of the processes of reasoned behaviours. Not only they help us understand reasoned behaviours, but are able to simulate these behaviours themselves. But when people tried to model the lower levels of behaviour, the artificial systems which were built failed to reproduce the behaviour observed in natural systems, mainly animals (Brooks, 1995). This was one of the strong reasons that motivated the development of BBS on the first place, but the fact is that BBS have modelled acceptably animal adaptive behaviour. BBS help us understand adaptive behaviour (*e.g.* Webb, 1996; 2001), but also we can build artificial systems which show this adaptiveness (Maes, 1991).

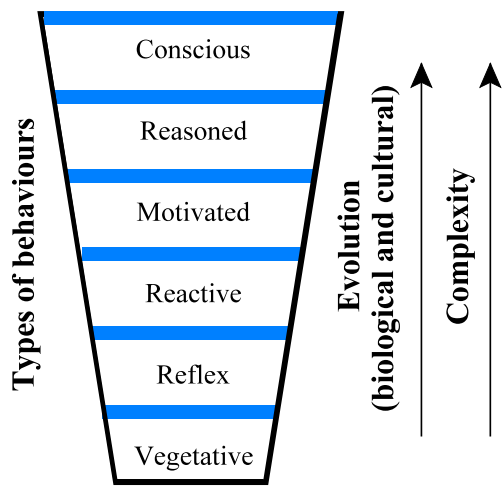

Figure 1. Abstraction levels in animal behaviour (Gershenson, 2001).

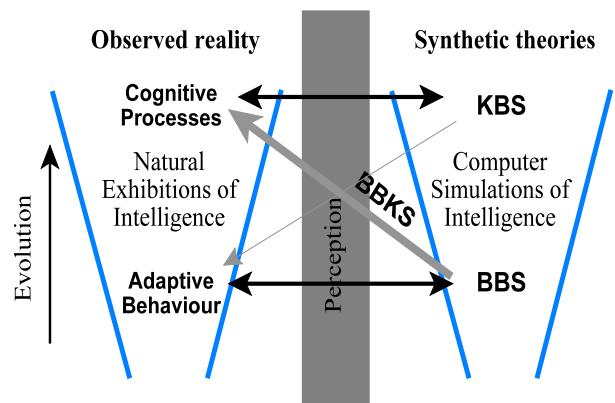

Figure 2. Simulating exhibitions of intelligence (Gershenson, 2001).

But if we believe that reasoned behaviours evolved and developed from lower levels of behaviour, we should start thinking how to justify this. Not only to validate our belief, but for understanding how was this possible, and to be able to reproduce it. *How can reasoning evolve and develop from adaptive behaviour?* We believe that this can be studied by simulating reasoned behaviours from a BBS perspective. This would not be a unification between BBS and KBS, but a bridge closing the gaps between them. They would be **behaviour-based knowledge systems** (BBKS). In other words, such a system should exhibit knowledge, which should have been *developed*, not directly implemented. In this way, a BBKS would be able to model the exhibitions of intelligence modelled by BBS and KBS, also illustrating the relationships between these types of intelligence: adaptive and cognitive. Also, BBKS are compatible with the Epigenetic Robotics approach (Balkenius *et. al.*, 2001; Zlatev, 2001).

We believe that this is a promising line of research. We argue that this is the most viable path for understanding most levels of behaviour, and therefore intelligence: natural and artificial. As the MacGregor-Lewis stratification of neuroscience notes, "the models which relate several strata (levels) are most broadly significant" (MacGregor, 1987, quoted in Cliff, 1991). Figure 2 shows a graphical representation of the ideas expressed above.

But is it possible to simulate knowledge from adaptive behaviour? We believe it is, and in the following section we describe how we might attempt to achieve this.

## 3. Knowledge from Behaviour

As we stated in the previous section, reasoned behaviours require abstract representations or concepts of the perceived world, and an accurate manipulation of these in order to produce a specific behaviour. How can these abstract representations and concepts be acquired? It seems that they are *learned* from regularities in the perceptions of objects and events. We believe that this is how concepts are created. We define a **concept** as a generalization of perception(s) or other concept(s)[1]. This definition requires and presupposes embodiment and situatedness (Clark, 1997). This means that if we intend to *simulate* these abstractions, our artificial creatures should be embodied and situated, or at least *virtually embodied and situated* (*i.e.* in simulations). We should note that in animals concepts are not physical structures (if we open a brain we will not find any concept): they emerge from the interactions of the nervous system with the rest of the body and environment. We can see them as a metaphor, and could say that they lie in the eye of the beholder. The same for other types of representation. And since these are necessary elements of knowledge, things will be clearer if we remark that knowledge is not a physical structure or element either, but an *emergent property* of a cognitive system (*i.e.* an observer needs to perceive the knowledge).

As an example for showing our use of concepts, a person begins to *develop* a concept of "pen" from the moment she perceives a pen. Then, when she perceives different instantiations of pens and their uses, all the regularities will determine her concept of "pen". We believe that animals also have such concepts and are shaped in the same way. A kitten might play with a ball of paper to explore what can be done with it. Once the kitten experiences the possibilities of sensation, perception and use, a concept representing the ball of paper should have been created, so that the animal will behave accordingly in future presentations of balls of paper. Of course, we have a different concept "ball of paper" than the kitten, because our perceptions (and the "hardware" we process them with) are different. But that a creature has different concepts than the ones we have, does not mean that it does not have concepts. The popular "problem" of the frog not having concept of a fly because frogs confuse other objects with flies (they try to eat them), is a bizarre anthropomorphization of the mind (based on the classical experiments by Lettvin *et. al.* (1959)). The frog has a concept, but not of a fly. Their perceptual system simply does not allow them to distinguish flies and similar objects. They do not need this to survive in their ecological niche. We observe a similar situation with fiddler crabs. They do not have a concept of "predator". They just have a concept of "something taller than me", and they run away from it (Layne, Land, and Zeil, 1997). This is because animals *develop* their intelligence to cope with *their* environment, not with ours. And even in humans, recent research (*e.g.* O'Regan and Noë, 2001; Clark, in press) shows that our visual perceptions are not as complete as they seem to us. We need to be aware of this while studying, and judging, animal and human intelligence. Concepts are necessary because it has a huge computational cost to remember each particular object and to act accordingly. Generalizations allow the cognitive system to produce similar actions in similar situations at a low computational cost.

But, strictly speaking, all humans also have different concepts for the same objects, since we have had different experiences of them. It is only because of **language** that we can communicate referring to the same classes of objects even when the mechanisms which determine in our brains those concepts might be very different from each other.

---

[1]This is not the classical notion of concept in the philosophy of mind literature (*e.g.* Peacocke, 1992), and it is not restricted to humans. It is compatible with the use of Gärdenfors (2000).

In animals, this ability to abstract concepts from perceptions should be given by the plasticity of the animal's neurons. This implies that some concepts might be innate, determined by the prenatal wiring of the neurons, which is dependent on the genome (and perhaps also proteome). How and why these innate mechanisms **evolve**, including the ones allowing the concept abstraction, are still open questions. One could argue that it is an advantage to have them, but recent studies (*e.g.* Alexander, 2001) have put a question on wether natural selection is the only (or even in some cases the main) driving force of natural evolution. Here we will not discuss this issue, just assuming that this ability has been already acquired. We will only say that work in *evolutionary robotics* (see Harvey *et. al.,* (1997) and Gomi (1998) for reviews) might lead to answers for these questions.

So, we can say that an animal is able to abstract regularities from its environment. We will not be aware of them if the animal does not exploit the acquired concepts in its behaviour. But if the animal *manipulates* the acquired concepts in order to *adapt* to its environment, we can say that the animal has abstracted *a logic* of its environment. *Knowledge of its environment*. As we stated, this is dependant on the observer, since we believe that knowledge is an emergent property of a cognitive system, not an element[2]. We can also say that the "proper" use of concepts gives them a certain meaning, grounded through action.

If we are searching for explanations of *our* logic, the logic of *our* environment, then we should take other issues into account. First of all, the fact that we live in a **society**[3], which is shaped by us, and shapes us. We have a **language**, which allows us to externalize and share our concepts. This allows us to have an access to the concepts of others, enlarging our knowledge. Language and human thought are so interrelated, interdependent, and internecessary, that some people even seem to have forgotten that they are different things[4]. We believe that language is also necessary for complex manipulation of concepts (Clark, 1998), and since an individual can develop a language only in a society (Steels and Kaplan, 2002), it is only in a society that an individual can develop higher cognition (Gershenson, 2001), as it seems has been in nature (Dunbar, 1998). Through generations, a **culture** is formed, accumulating past experiences.

Summing up, the epigenetic stages we should follow to reach knowledge from behaviour, should be:
1. concept abstraction.
2. grounding of concepts though action.
3. sharing of concepts through social interactions (language).
4. manipulation of concepts (logic[5]).
5. evolution of concepts (culture).

Note that knowledge is not acquired only until completing all the stages, but it is *developed gradually* with every stage. And we will not say that our knowledge cannot be improved as well, *i.e.* we can always add more stages. Also, steps 3 and 4 could exchange places. In fact, there have been models and theories which address most of the steps described above, but separately (*e.g.* Scheier and Lambrinos, 1996; Gärdenfors, 2000; Zlatev, 2001; Cangelosi and Parisi, 2001; Prince, 2001; Steels and Kaplan, 2002), and thus they answer only partially the question of how knowledge could evolve from adaptive behaviour. Of course, it is necessary to have such models and theories before attempting to model all the path, and they all can be considered as BBKS or BBKS theories.

These requirements for acquiring higher order cognition seem quite sensible, and have been proposed with similar approaches (*e.g.* Kirsh, 1991; Steels, 1996; Clark, 1997; Balkenius *et. al.*, 2001; Zlatev 2001; Gershenson, 2001; Steels and Kaplan, 2002).

Another way of convincing ourselves to follow this path is to analyze it backwards: if we take from humans each of the stages described, how do our knowledge would be diminished? Without culture we would not be able to accumulate knowledge from generation to generation, and only by our physiological abilities we would be at a level even lower than social primates. Without being able to manipulate concepts we would not be able to make inferences nor predictions. Without a language we would not be able to learn what other individuals have learned, and we would be restricted to our individual learning. Without concepts, we would just have reactive behaviours, without the possibility of integrating our sensory experiences to produce complex behaviours. But of course, for using these concepts they need to be grounded.

We believe that following this approach, consistent with Epigenetic Robotics (Balkenius *et. al.*, 2001; Zlatev, 2001), we will be able to build systems which *develop* their own logic, consistent with *their* environment, which will be able to do reasonings in the sense a KBS does. Of course, this will not replace KBS, since their manageability at a knowledge level is much higher than it would be in BBKS (as noted in

---

[2] We should also be careful with language games, such as "Does a tree *knows* when spring came because it blossoms?".

[3] The social factor has been proposed to be also responsible for the evolution of our "big" brains (Dunbar, 1998).

[4] Though thought seems to have all the properties of a language...

[5] We mean logic as a tool, not logic as a science.

Gärdenfors (2000)). But KBS deliver us no knowledge of *how knowledge takes place*, whereas this is the goal of BBKS.

## 4. Limits of BBKS

The proposed line of research is no panacea. We can already see several limitations of this approach, which should be considered if to follow this path.

When we model natural exhibitions of intelligence (Figure 2), some people might say that we sin of oversimplification, because we do not model all the conditions which affect a natural cognitive system. But simplification is a necessity, due to the immense complexity of the phenomena which are modelled. It is this complexity, and all the information (even when it might be redundant) that our cells and brains can contain, that force us to make simplifications in our models. We believe that this information is so huge that the complexity of natural organisms exhibiting intelligent behaviour cannot be simulated in artificial systems without simplification[6]. Our actual computers are very far from being able to calculate in real time all the necessary operations which a realistic (non oversimplifying) model would require. Some alternatives might lie in DNA computing (Benenson *et. al.*, 2001), but even if we had such computational power as the one required to imitate convincingly living organisms, how to program all the necessary information? At this moment this seems impossible in a short time scale.

But where to go? It seems we can make a distinction depending on our purposes. If we are interested in *understanding* intelligence (as we are with the development of BBKS), then our limited models creating artificial systems seem to suit our purposes. If we want to *produce* intelligence "higher than human", we can learn a bit from the history of such attempts. In the beginnings of Artificial Intelligence, some people assumed that all the knowledge of a human adult might be *programmed*. Other people aware of obvious difficulties, looking at how natural systems acquire their knowledge, thought of programming a "child" computer that would be able to *learn* as a child does (*e.g.* Turing, 1950) (of course, "one could not send the machine to school without the other children making excessive fun of it" (Turing, 1950)), (it is easier if you do not program everything but let the parameters be adjusted by the system, *i.e.* learned). Another alternative has been to *evolve* the mechanisms in charge of producing intelligent behaviour, also being inspired in nature (it is easier if you do not program everything but let the model be adjusted by the system, *i.e.* evolved), but it has required too much computational power in order to aspire to reach "higher order" intelligence by itself. These alternatives and combinations of them have been used depending on the ideas and purposes of researchers, modelling from bacteria to human societies, all of them "sinning of oversimplification". So, if we want to produce intelligence "higher than human", it seems sensible that we should not start building the computational mechanisms from scratch. This is, we should not attempt to throw away five billion years of evolution and the computational power of our cells, and start from where we already are, even from the hardware perspective. This implies that we should build our systems *on us* (The best model of a cat is another cat, and if possible, the same cat). But for this of course we need first to understand with our limited simplified models how our mind works, in order to try to improve it.

---

[6] This idea is clearly presented by Michael Arbib (1989), speaking about brain models: "a model that simply duplicates the brain is no more illuminating than the brain itself" (p. 8).

But we should notice that we already make intelligence "higher than human", just with our cultural and technological evolution. There was already human intelligence more than two thousand years ago, but we could say that we are able to exhibit more intelligence (we are able to solve more tasks) than humans of even a hundred years ago (*e.g.* you can make calculations much easier with a computer than with pen and paper). By altering the nature of our environments, changing them to suit our purposes, we make our environments and our tools to manipulate them more complex, and our intelligence can be considered to be higher (Clark, 2003). Or from another perspective, we raise the level of human intelligence, even with roughly the same "hardware" (Our DNA has not changed much in the last ten thousand years). Cultural evolution implies that the intelligence will be improved each generation. And the understanding of this process, will allow us to guide it.

## 5. A Behaviour-based Knowledge System

We are currently developing a BBKS in order to study the development of knowledge in artificial cognitive systems. Following the ideas presented in Gershenson, González and Negrete (2000), we are constructing a virtual laboratory in order to contrast our models as virtual animats develop and survive in their environment. This virtual laboratory can be downloaded (source code included) from http://www.cogs.sussex.ac.uk/users/carlos/keb. A screenshot of the virtual environment can be appreciated in Figure 3.

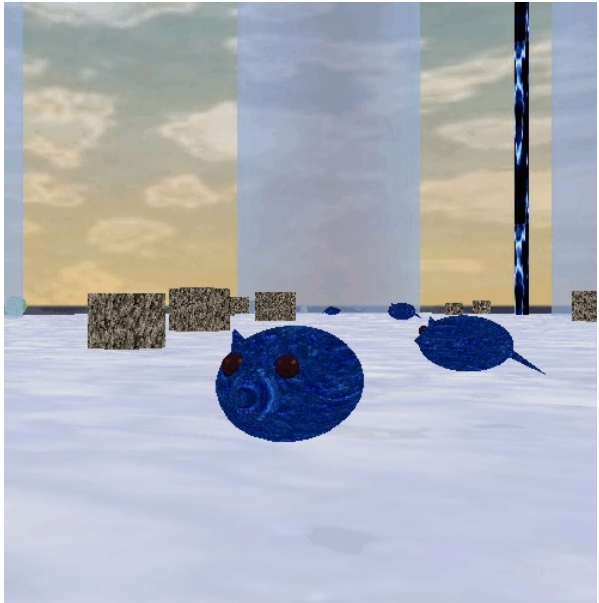

Figure 3. Screenshot of the virtual environment.

At this stage, animats are able to extract regularities from their sensors developing data structures which we call "koncepts". Then regularities in these koncepts recursively form koncepts of a higher level. With a simple reinforcement learning scheme, koncepts are linked to actions, grounding a form of "meaning" of the koncepts. Using these koncepts for select their actions, animats are able to survive in their simple environment.

We are currently developing more complex tasks, in order to model how logic can emerge from the use of abstract koncepts, in a similar way as the one proposed by Gärdenfors (1994). We are also interested in studying the sharing of koncepts through a form of communication, and how this affects the cognitive development of the animats. An extensive description of this work will be found in Gershenson (2002b).

## 6. Conclusions

We have proposed the development of behaviour-based knowledge systems for explaining the transition of from adaptive behaviour to high cognitive processes in a synthetic fashion. This is, with our artificial systems we are not only understanding the natural systems which inspire us, but at the same time we become capable of engineering systems with the potentialities of the natural ones. We have stated broadly the steps and requirements that BBKS should follow for producing knowledge of their environment while still exhibiting adaptive behaviour. We have also discussed some limitations of BBKS and presented briefly our current work.

An additional motivation for developing BBKS is for doing philosophy of mind and philosophy of cognitive science with the aid of synthetic systems: *synthetic philosophy*. In this way, theories of mind, concepts, meaning, representation, intentionality, consciousness, etc. could be contrasted with our synthetic BBKS, reducing a bit the space for rhetoric.

## Acknowledgements

I appreciate the valuable comments and suggestions received from Nadia Gershenson, Ezequiel Di Paolo, Inman Harvey, Andy Clark, David Young, Douglas Hofstadter, Peter Gärdenfors, Christian Balkenius, the cognitive science group at Lund University, and two anonymous referees. The main ideas were derived from the research carried out with Pedro Pablo González and José Negrete. This work was supported in part by the Consejo Nacional de Ciencia y Tecnología (CONACYT) of México and by the School of Cognitive and Computer Sciences of the University of Sussex.